\documentclass[journal]{IEEEtran}
\usepackage{times}
\usepackage{epsfig}
\usepackage{graphicx}
\usepackage{amssymb}
\usepackage{xcolor}
\usepackage{booktabs}
\usepackage{multirow}
\usepackage{subfigure}

\usepackage{xspace}

\usepackage[pagebackref=true,breaklinks=true,letterpaper=true,colorlinks,bookmarks=true]{hyperref}

\ifCLASSOPTIONcompsoc
  \usepackage[nocompress]{cite}
\else
  \usepackage{cite}
\fi

\usepackage{amsmath}
\makeatletter
\DeclareRobustCommand\onedot{\futurelet\@let@token\@onedot}
\def\@onedot{\ifx\@let@token.\else.\null\fi\xspace}

\def\eg{\emph{e.g}\onedot} 
\def\ie{\emph{i.e}\onedot} 
\def\cf{\emph{c.f}\onedot} 
\def\etc{\emph{etc}\onedot} 
 
\def\etal{\emph{et al}\onedot}
\makeatother

\usepackage{algorithmic}

\usepackage{array}

\usepackage{stfloats}
\usepackage{url}

\begin{document}
%
\title{Deepfake Forensics via An Adversarial Game}
%
%
%
%

\author{Zhi~Wang,
        Yiwen~Guo,
        and~Wangmeng~Zuo
\thanks{Y. Guo was with ByteDance AI Lab. E-mail: guoyiwen.ai@bytedance.com.}
\thanks{Z. Wang and W. Zuo are with the School of Computer Science and Technology, Harbin Institute of Technology, Harbin 150001, China. E-mail: cszwang651@gmail.com, cswmzuo@gmail.com.}
\thanks{Manuscript received March 1, 2021.}}

%
%

\markboth{SUBMISSION TO IEEE TRANSACTIONS ON IMAGE PROCESSIN}%
{}
%



\IEEEtitleabstractindextext{%
\begin{abstract}
With the progress in AI-based facial forgery (\ie., deepfake), people are concerned about its abuse. Albeit effort has been made for training models to recognize such forgeries, existing models suffer from poor generalization to unseen forgery technologies and high sensitivity to changes in image/video quality. In this paper, we advocate robust training for improving the generalization ability. We believe training with samples that are adversarially crafted to attack the classification models improves the generalization ability considerably. Considering that AI-based face manipulation often leads to high-frequency artifacts that can be easily spotted (by models) yet difficult to generalize, we further propose a new adversarial training method that attempts to blur out these artifacts, by introducing pixel-wise Gaussian blurring. Plenty of empirical evidence show that, with adversarial training, models are forced to learn more discriminative and generalizable features. Our code: \url{https://github.com/ah651/deepfake_adv}.
\end{abstract}

\begin{IEEEkeywords}
Deepfake forensics, adversarial training, data augmentation, generalization ability.
\end{IEEEkeywords}}

\maketitle

\IEEEdisplaynontitleabstractindextext

%
\IEEEpeerreviewmaketitle

\section{Introduction}\label{sec:introduction}

\IEEEPARstart{T}{he} rapid development of deep learning and generative modeling have promoted the progress of face manipulation and forgery technologies (\ie., deepfake), which largely lower the bar of creating photo-realistic facial images/videos. 
At present, there exist a variety of AI-based methods, which can be used for identity swapping, expression editing, attribute manipulation, \etc. Face2Face~\cite{face2face}, for instance, is a method that can edit the expression of a person to make it identical to that of another person. Face replacement methods like FaceSwap~\cite{faceswap_project} is capable of further replacing the facial region of somebody with that of another person in a video. 

The counterfeit products of AI-based facial manipulation or forgery make it difficult for human beings and conventional facial classification systems to distinguish between real and fake. Deepfake technologies are thus at high risk of malicious abuse, posing threats to face recognition applications, \eg., face recognition based payment and access control. Malicious facial manipulation and forgery may also infringe on individual privacy and reputation. Hence, in order to protect public safety and individual privacy, it is essential to develop methods for detecting deepfake.

In this work, we focus on the problem of detecting facial forgeries, that is, automatically detecting whether an image (or a video) of a human face has been manipulated or was forged by AI-based technologies. We adopt the term ``deepfake detection" to describe the task, following prior arts. 

A variety of methods have been developed for deepfake detection. Early work relies on handcrafted features, using for example noise variance analysis~\cite{pan2012exposing} and digital shadow writing analysis~\cite{cozzolino2014image, fridrich2012rich} to discover differences between real and synthetic images/videos. Recently, deep-learning-based deepfake detection methods have been proposed and discussed, which advocate to use deep models like deep convolutional neural networks (CNNs) to achieve the goal. 
Though significant technical progress has been made over the last few years, existing methods normally suffer from generalizing to images/videos generated by unseen technologies or even unseen models~\cite{yu2019attributing}.
That is, they can achieve a very high accuracy of $\sim98$\% for a manipulation technology whose output images/videos have been seen during training of the detection model, yet fail to generalize to images/videos generated by unseen technologies.

There are already several methods trying to resolve this overfitting problem and improve the performance of deepfake detection, but most of them, \eg,~\cite{face-xray, du2019towards}, rely on modifying the architecture of the classification models. 
In this paper, we attempt to address the problem from a different perspective, by innovating the training mechanism of deepfake detection models and adopting adversarial training.
In general, adversarial training aims at discovering challenging samples that are not easily predicted by the current classification model, and we believe training with these samples encourages the model to focus on more essential and generalizable features that could be used to distinguish the evolving fake images/videos from the real ones, thereby improving the model's generalization ability to unseen forgeries. 

We evaluate several different types of adversarial examples in adversarial training, including the additive~\cite{goodfellow2014explaining} and spatial-transformed adversarial examples~\cite{stadv}. We also propose adversarially blurred examples which can be more suitable to the task, leading to improved generalization performance to both unseen forgery technologies and unseen image/video qualities in the test phase.
Except for the input-gradient-based strategy for crafting adversarial examples~\cite{goodfellow2014explaining, madry2017towards}, a generator-based strategy is also advocated, which not only controls the computational cost for obtaining adversarial examples, but also improves their flexibility. 
Extensive experimental results verifies the effectiveness of our method.
In summary, our contributions are:
\begin{itemize}
\item We introduce adversarial training into the training process of deepfake detection models. We show that it improves the generalization ability and robustness of the models notably.
%
\item A novel method of generating adversarial examples based on image blurring is proposed, and it is shown to be more suitable to the adversarial training framework of deepfake detection.
\item Since our method focuses on innovating training strategies, our proposed adversarial training framework can be used together with many existing methods which modify the network structure to further improve the performance of deepfake detection model.
%
\item Extensive experiments show that the performance of several popular deepfake detection models can be improved by using our method, in the sense of better generalization ability on unseen forgery technologies and image/video qualities. 
\end{itemize}


\section{Related Work}

\textbf{Deepfake Generation.}
Research on AI-based face manipulation and forgery technologies has a long history. Although early work shows pleasing results only in very restricted scenarios~\cite{bitouk2008face, bregler1997video}, over the last decade, with the rapid development of computer graphics and computer vision, facial manipulation has become more and more photo-realistic. 
For instance, Dale \etal~\cite{dale2011video} managed to swap faces in videos by reconstructing 3D face models of different people. 
3D-based methods were also used by Garrido \etal~\cite{garrido2014automatic} and Thies \etal~\cite{face2face}, sometimes in combination with neural rendering technologies~\cite{neuraltexture}. 
In addition to the graphics-based methods, there are also many vision-based methods.
In particular, the recent upsurge of deep learning has made these methods (\eg, DeepFakes~\cite{deepfake}, FaceSwap~\cite{faceswap_project}, and ZAO~\cite{ZAO}) popular for synthesizing photo-realistic facial images, and the term ``deepfake'' also comes from this trend.
Generative adversarial nets (GANs)~\cite{goodfellow2014GAN} can also be used for facial attribute editing~\cite{stgan, choi2018stargan, he2019attgan}, face swapping~\cite{nirkin2019fsgan, bao2018towards, olszewski2017realistic}, etc. In addition, GANs~\cite{goodfellow2014GAN} were also used for direct synthesis of whole facial images from noises.

\textbf{Deepfake Forensics.}
Maliciously forged deepfake images/videos are harmful to individual privacy, and they can pose a grave threat to the society. It is thus of great importance to develop deepfake detection solutions. While early attempts~\cite{pan2012exposing, fridrich2012rich, goljan2015cfa, ferrara2012image} focused on internal statistics or handcrafted features of images/videos to discover the difference between real and fake, most recent methods were usually based on deep learning features~\cite{zhou2017two, cozzolino2017recasting, rahmouni2017distinguishing, bayar2016deep, li2018exposing} or end-to-end trained deep binary classifiers~\cite{afchar2018mesonet, rossler2019faceforensics++, guera2018deepfake, stehouwer2019detection, zhao2021multiattentional}. 
As have been criticized, most of the methods suffer from severe overfitting to the training data and cannot be effectively used in many practical scenarios. 
There are methods trying to cope with the overfitting issue of deepfake detection models. For instance, Li 
\etal~\cite{face-xray} proposed a more generalizable deep face representation for achieving the goal, by introducing an auxiliary task of predicting face X-ray (\ie, artifacts incurred by blending fake faces into the original images). Stehouwer \etal~\cite{stehouwer2019detection} modified the network architecture for deepfake detection and introduced an attention module for the task.
Zhao \etal~\cite{zhao2021multiattentional} formulated deepfake detection as a fine-grained classification problem and proposed a new multi-attentional deepfake detection network. Auto-encoders were also considered~\cite{du2019towards}. 
Sun \etal~\cite{sun2021improving} detected deepfake videos through temporal modeling on precise geometric features.
In addition, it is also a common idea to use multi-modality to improve the robustness of the model. Zhou \etal~\cite{zhou2021joint} exploited intrinsic synchronization between the visual and auditory modalities to facilitate deepfake detection.
Unlike these methods that innovated network architectures, our work in this paper focus solely on the training mechanism of deepfake detection models, thus it is orthogonal to these efforts and can be naturally combined with them to achieve even better results.

\textbf{Adversarial Training.}
Adversarial training refers to a training mechanism that utilizes adversarial examples for augmenting the training set, which constitutes the main basis of defense against adversarial attacks~\cite{goodfellow2014explaining, kurakin2016adversarial, madry2017towards, uesato2019labels}. The development of adversarial training can be traced back to~\cite{goodfellow2014explaining}, in which the fast gradient sign method (FGSM) was proposed for improving the adversarial robustness. Madry \etal~\cite{madry2017towards} further proposed to use a multi-step scheme called projected gradient descent (PGD), enabling more powerful robustness than that obtained with FGSM and many of its contemporary defense methods~\cite{athalye2018obfuscated}. 
Hussain \etal~\cite{hussain2021adversarial} revealed the vulnerability of existing deepfake detection models to adversarial examples. Ruiz \etal~\cite{ruiz2020disrupting} adopted the method of generating adversarial examples to prevent photos from being used for generating deepfakes. 
In contrast to their methods, we advocate adversarial training for improving the performance of the classification-based deepfake detection model.
In Sec.~\ref{sec:approach}, we will discuss how adversarial training can benefit the performance of deepfake detection, and we will also introduce a new type of adversarial examples which is more suitable to the deepfake detection task. 

\section{Proposed Approach}\label{sec:approach}
This section introduces our proposed framework (based on adversarial training) for deepfake detection. First, we explain why adversarial training is advocated, and we will also revisit the basic concept of adversarial training, mostly on the basis of commonly used additive adversarial examples in Sec.~\ref{sec:adv_train}. Then, in Sec.~\ref{sec:blur_adv}, we introduce a new and dedicated method for performing adversarial attacks and adversarial training, which is based on pixel-level Gaussian blurring. 
Finally, we introduce how generator-based adversarial training can be performed in Sec.~\ref{sec:generator}. 

\begin{figure*}
    \begin{center}
        \includegraphics[scale=0.17]{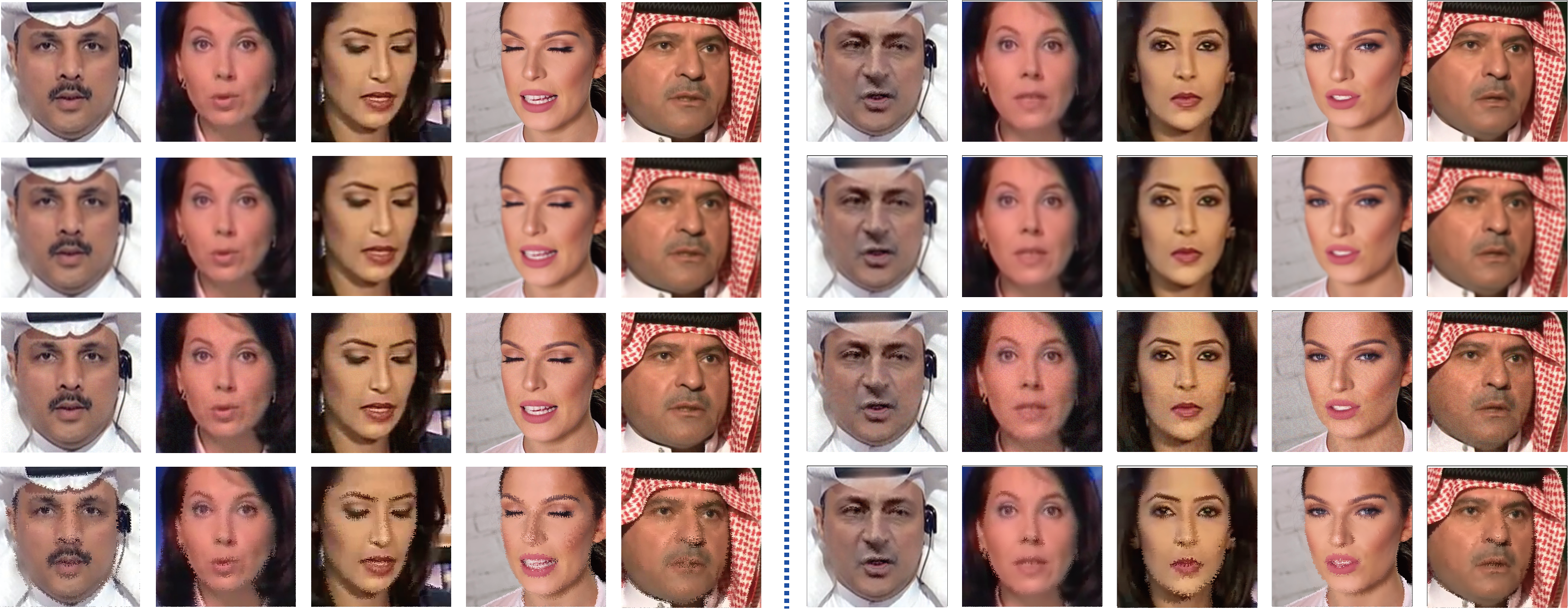}
    \end{center}
    \vskip -0.15in
    \caption{Visualization results of various adversarial examples. From the top row to the last, we show results of the original images, our adversarially blurred examples, the FGSM examples~\cite{goodfellow2014explaining}, and the spatial-transformed adversarial examples~\cite{stadv}. On the left half, we show adversarial examples crafted on real face images, and the right half we show those crafted on forged or manipulated faces.}
    \label{fig:adv_sample}
\end{figure*}

\subsection{Adversarial Training Framework}\label{sec:adv_train}
Deepfake detection is normally cast into a binary classification task. Predictions can be made on the basis of one image (as model input) or a sequence of images in a single video. For the sake of simplicity, here we consider the case where only an image is fed to the model, and we note that our method can naturally generalize to models whose inputs are a sequence of images~\footnote{See the last paragraph of Sec.~\ref{sec:adv_exps} for empirical evidence.}. Given a training set $\mathbb{D}$ that includes a large number of images and their corresponding labels. One limitation of many existing deepfake detection models is that normal training on $\mathbb{D}$ barely guarantees the generalization to fake images generated by unseen technologies or compressed with different quality factors. A plausible solution to the problem is to introduce an ``adversary'' that keeps refining the training fake images and removing obvious artifacts that could easily be spotted by the deepfake detection model, such that the model learns to correctly classify more advanced fake images.
This is in coordinate with the spirit of adversarial learning. 

Let us revisit the traditional adversarial learning formulation first. Assume that a classification model (\eg, the deepfake detection model) attempts to minimize the prediction loss $L(x, y; \theta)$ for any given data $(x,y)$ (\ie, an image $x$ paired with its label $y$), in which $\theta$ collects all learnable parameters in the classification model, $x \in \mathbb R^{h\times w\times c}$ ($h$, $w$, and $c$ represent height, width, and number of channels of $x$, respectively), and $y\in\{0,1\}$. The goal of the normal training mechanism is to find an appropriate set of parameters to minimize the empirical risk $\sum_{(x, y) \in \mathbb{D}}[L(x, y; \theta)]$, while, aiming at addressing the adversarial vulnerability of deep models, adversarial training strengthens the models by generating adversarial examples and injecting them into the training set. 

Over the past few years, a variety of methods of generating adversarial examples have been proposed, and the most popular methods add pixel-level perturbations to natural images, \eg, FGSM~\cite{goodfellow2014explaining} suggests to obtain each adversarial example by adding a scaled input-gradient sign to each natural image $x$ in a set. That is:
\begin{equation}
    \label{equ:fgsm}
    x^{adv} = x + \epsilon\cdot \text{sign} (\nabla_x L(x, y; \theta)) .
\end{equation}
The above equation is derived from an optimization problem maximizing the prediction loss of an input obtained by adding a perturbation (whose $l_\infty$ norm is no greater than $\epsilon$)~\cite{goodfellow2014explaining}. 

Adversarial training plays a zero-sum game which includes an auxiliary process that generates adversarial examples which maximizes the classification loss. The generated adversarial examples can be used instead of the original benign examples or in combination with them for training. 
For the former, we have
\begin{equation}
    \label{equ:cls_loss}
    \min_\theta
    \sum_{(x,y) \in \mathbb{D}}
    \max_{\delta \in \mathbb{S}}\ L(x+\delta, y; \theta) ,
\end{equation}
while for the latter, we can use the following optimization problem instead of~(\ref{equ:cls_loss}):
\begin{equation}
    \label{equ:cls_loss2}
    \min_\theta
    \sum_{(x,y) \in \mathbb{D}}
    L(x, y; \theta) + \lambda\cdot
    \max_{\delta \in \mathbb{S}}\ L(x+\delta, y; \theta) .
\end{equation}
Note that we introduce a set $\mathbb{S} \subseteq \mathbb{R}^{h\times w\times c}$ to constrain the allowable disturbance from each adversarial example to its corresponding natural (or say ``clean'') image. For FGSM as introduced in Eq.~(\ref{equ:fgsm}), we have $\mathbb S=\{z|\|z\|_\infty\leq \epsilon\}$. In general, $\mathbb{S}$ guarantees the visual similarity between the adversarial example and the natural image.

\begin{figure*}
    \begin{center}
        \includegraphics[scale=0.163]{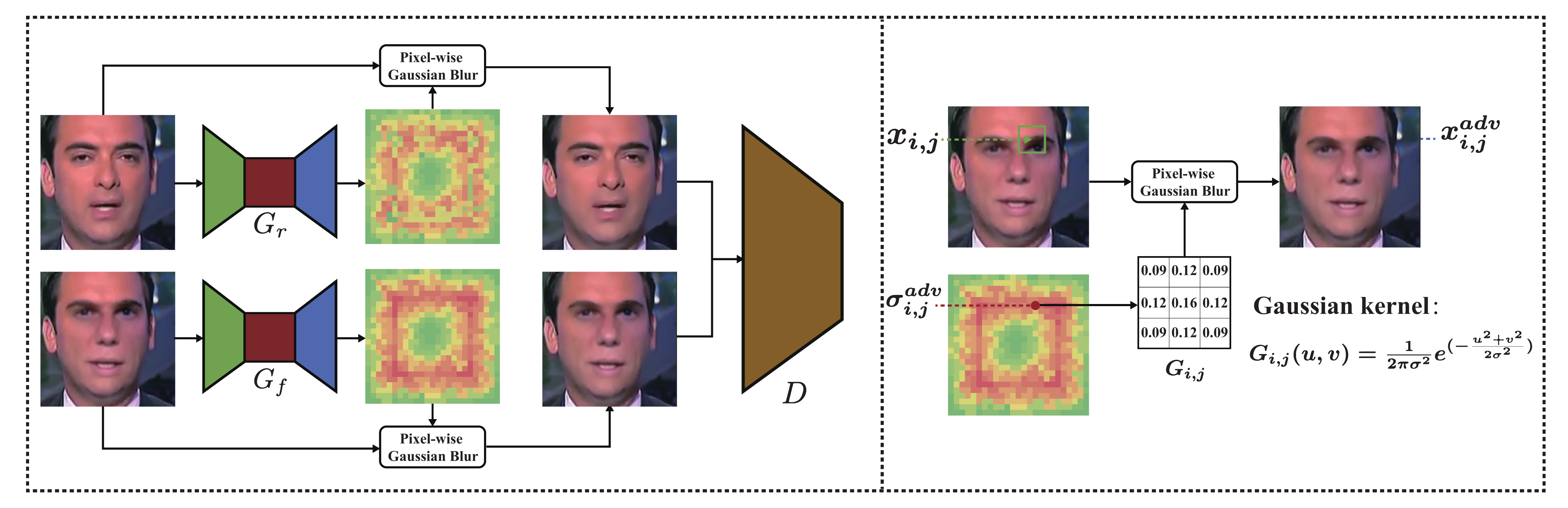}
    \end{center}\vskip -0.15in
    \caption{An overview of our two-generator-based blurring adversarial training (Two-Gen-BAT). The right panel shows how the adversarial blurring is performed. For each pixel $x_{i,j}$ on the original image, we use the corresponding $\sigma_{i,j}^{adv}$ to generate a Gaussian kernel $G_{i,j}$, and then use $G_{i,j}$ to perform pixel-wise Gaussian blurring on $x_{i,j}$ by considering its surrounding pixels to obtain a pixel of the adversarial example, \ie, $x_{i,j}^{adv}$. The $\sigma$ map was resized for better visualization.}
    \label{fig:overview} \vskip -0.04in
\end{figure*}

Recently, several other kinds of adversarial examples have also been proposed. 
For instance, instead of imposing pixel-level additive perturbations, Xiao \etal~\cite{stadv} proposed to calculate an adversarial optical flow to spatially transform each pixel of the natural images accordingly.
Let us use $x_{i,j}$ to represent the pixel on the $i$-th row and $j$-th column of the original clean image, an adversarial optical flow $f_{i,j} := (\triangle u_{i,j}, \triangle v_{i,j})$ is learned in the method to transform $x_{i,j}$ to the corresponding position $(i',j') = (i + \triangle u_{i,j},j + \triangle v_{i,j})$ on the adversarial image $x^{adv}$. The magnitude of the adversarial optical flow is encouraged to be small while leading to large prediction loss of the adversarial example. 
Training with this sort of examples will be called spatial-transformed adversarial training (SAT), and training with the result of Eq.~(\ref{equ:fgsm}) will be called additive adversarial training (AAT) in this paper.
Besides the introduced ones, there exist other types of adversarial examples, \eg, those utilizes white-balance~\cite{afifi2019else}.

\subsection{Blurring Adversarial Training}\label{sec:blur_adv}

Albeit adversarial training based on the aforementioned examples have achieved improved robustness under adversarial attacks, their performance in enhancing the generalization ability of deepfake detection models is unclear. In fact, on natural image classification tasks (\eg, on ImageNet~\cite{imagenet}), it has been demonstrated that adversarial training barely contributes to the generalization ability to normal test data, on account of the distribution drift between these adversarial examples and normal test samples, and the same problem might also exist in the task of deepfake detection.
Here we propose a new type of adversarial examples that is shown to be more effective in the adversarial training framework for deepfake detection.

We know from prior work~\cite{wang2020cnn} that some high-frequency components of the fake images/videos are very easily spotted by models yet also very specific and difficult to generalize. 
That is, introducing Gaussian blurring and JPEG compression~\cite{wang2020cnn} augmentations probably improves the deep classification CNNs, and it might be even more effective to introduce a blurring-based adversarial training mechanism.
Specifically, given an input image $x$ whose height, width, and number of channels are $h$, $w$, and $c$, respectively, we obtain $x^{adv}$ by performing pixel-wise Gaussian blurring on $x$. We use $x_{i,j}^{adv}$ to represent the $(i,j)$-th pixel of $x^{adv}$, and we attempt to learn a single-channel map $\sigma^{adv}$ with a size of $h\times w$, each of whose entries (\eg, $\sigma^{adv}_{i,j}$) represents the standard deviation of a Gaussian kernel to be applied to the region centered at the corresponding pixel of image $x$, \ie, $x_{i,j}$. In details, for obtaining the value of $x^{adv}_{i,j}$, we first collect $\sigma^{adv}_{i,j}$, then use it to calculate the kernel $G_{i,j}\in\mathbb R^{k\times k}$ for performing Gaussian blurring around $x_{i,j}$. Suppose that the kernel size is chosen as $k$, then we calculate the inner product between $G_{i,j}$ and $\gamma(x_{i,j}, k)\in\mathbb R^{k\times k}$ (\ie, a region of pixels centered at the pixel $x_{i,j}$ with a radius of $k$).
That is:
\begin{equation}
    G_{i,j}(u, v) = \frac{1}{2 \pi (\sigma^{adv}_{i,j})^2} \exp \left(- \frac{u^2 + v^2}{2 (\sigma^{adv}_{i,j})^2}\right) ,
\end{equation}
\begin{equation}\label{equ:blur_adv}
    x^{adv}_{i,j} = \langle G_{i,j}, \gamma (x_{i,j}, k) \rangle ,
\end{equation}
in which $u$ and $v$ represent the relative coordinates to the centre pixel in $\gamma(x_{i,j}, k)$.
Such a pixel-wise Gaussian blurring can be easily implemented as a vectorized operation and thus is computationally very efficient. The map $\sigma^{adv}$ basically controls how much blurring is to be performed on the original training image. Larger entries of $\sigma^{adv}$ should lead to more blurry images and leave less obvious artifacts from the deepfake generator, while on the contrary, smaller entries of $\sigma^{adv}$ leave more obvious artifacts for the classification model to learn. We have $x^{adv}\rightarrow x$, as all the entries of $\sigma^{adv}$ approach zero.

We aim at learning a reasonable map $\sigma^{adv}$ for each training image. Since the adversarial blurring is to be performed in a pixel-wise manner, we are able to blur more on image regions with less generalizable features.
Similar to other adversarial examples, here the blurring-based adversarial examples are suggested to have less distortions from the original images, and we introduce a simple one-step scheme to achieve the goal, just like FGSM (except for the sign function): 
\begin{equation}\label{equ:one_step}
    \sigma^{adv} = \sigma + \epsilon \cdot \nabla_\sigma L(x^{adv}, y; \theta),
\end{equation}
in which $x^{adv}$ is obtained by Eq.~(\ref{equ:blur_adv}) and $\sigma$ is an initialization of $\sigma^{adv}$. In practice, we let $\sigma$ be a matrix whose entries share a common value (\eg, $1$). Some adversarially blurred examples are illustrated in Fig.~\ref{fig:adv_sample}, showing that they look as natural as the original images. 

One can further extend the simple one-step scheme in Eq.~(\ref{equ:one_step}) to a multi-step scheme, just like from FGSM~\cite{goodfellow2014explaining} to PGD~\cite{madry2017towards}. This requires iteratively performing Eq.~(\ref{equ:one_step}), and it can indeed help achieve higher attack success rates. However, it also leads to longer training time in our adversarial training framework. Adversarial training can further be adopted on the crafted examples. We will call this method blurring adversarial training (BAT).

\subsection{Generator-based Methods}\label{sec:generator}
Most adversarial training methods craft adversarial examples using the input-gradient of the loss function $L$. As has been mentioned, powerful adversarial examples are normally designed with a multi-step scheme, therefore the computational cost increases as the the number of steps increases. We propose an alternative way of generating adversarial examples, by introducing a CNN-based generator to control the training complexity, somewhat similar to~\cite{rusak2020simple}. Assume there are $M$ neurons in each layer of a $N$-layer MLP deepfake detection model, then it requires $O((w \times h \times c + 2) \times M + (N-2) \times M^2)$ floating point operations for performing the forward and backward pass each for computing the gradient in Eq.~(\ref{equ:one_step}), and using a $K$-step scheme makes the computational complexity of generating an adversarial example be $O(2\times(w \times h \times c + 2)\times K\times M + 2\times(N-2) \times K\times M^2)$, which is obviously related to the architecture of the deepfake detection model. While with the adversarial generator incorporated, they can be uncorrelated and solely depends on the architecture of the generator. That is, we can easily constraint the complexity by controlling the size of the generator.

We consider a CycleGAN~\cite{zhu2017unpaired} generator for generating $\sigma^{adv}$. We emphasize that, unlike the work by Rusak \etal~\cite{rusak2020simple}, for each original training sample $(x,y)$, we generate a specific map $\sigma_{(x,y)}^{adv}$ (the subscript $(x,y)$ will be omitted in the paper though) for it, considering the fact that the most transferable features on different images can reside in different spatial regions. Denote by $\theta_D$ and $\theta_G$ the set of learnable parameters for the deepfake detection model and that for the generator, respectively, we opt to playing the following min-max game:
\begin{equation}
    \label{equ:min_max}
    \min_{\theta_D} \max_{\theta_G} \sum_{(x, y) \in \mathbb{D}} L(x, y; \theta_D) 
    + \sum_{(x, y) \in \mathbb{D}} L(G(x; \theta_G), y; \theta_D) ,
\end{equation}

The introduced generator $G$ can also be considered as an enhancement model for the original deepfake generator(s).
The optimization problem Eq.~(\ref{equ:min_max}) allows to train a generator $G$ whose goal is opposite to the deepfake detection model, \ie, to remove obvious artifacts and synthesize more realistic deepfakes that can invalidate the deepfake detection model. If the generator indeed learns to synthesize more realistic fake images, then the classification model can learn more about deepfake and thus becomes more generalizable. The generator in turn learns to further improve its generation ability. More importantly, the generator-based method can be more flexible, and it suffices to learn to conceal more generalizable features in different images in combination with BAT, as will be shown in our experiments. 
In practice, our generator is used to ``enhance'' both fake and real training images, to balance training data from both classes.

Our generator-based BAT is also related to GAN~\cite{goodfellow2014GAN}, which contains a pair of generator and discriminator as well. What makes our method inherently different is that: the goal of our framework is to improve the deepfake detection model (\ie, our discriminator), while GAN aims to improve its generator.
In our case, the generator is used to craft adversarial examples, while the goal of a GAN generator is to capture the distribution of natural images.
Our discriminator is used to distinguish fake from real images (all ``enhanced'' by the generator), while the GAN discriminator is used to distinguish whether an image is a synthesized one (synthesized by the GAN generator) or a natural image (directly collected from the training dataset).
An optimal generator in our framework should be capable of removing all artifacts in deepfake contents or adding similar artifacts to real images to fool the discriminator, however, such a min-max game suffers from convergence to the Nash equilibria~\cite{farnia2020gans,mescheder2017numerics}, and thus the obtained generator may only be able to manipulate obvious and less generalizable artifacts that are easily captured by the deepfake detection model.

\textbf {Two Generators.}
Since the distribution of the real images and that of the fake images are different, we might need to introduce a very large generator to adapt it to both the two classes. Furthermore, the generator will have to first predict whether its input is real or fake and then attempt to process it to make it more like a fake one or a real one, to achieve the aforementioned goal.
On this point, we propose to use two generators for images from the two classes, respectively, to alleviate this problem. That is, we introduce $G_{{r}}$ which only processes real images and $G_{{f}}$ which only processes fake images. By introducing the two different generators, each of them will only be responsible for images from a single class, and we can expect them to learn more specific adversarial strategies for the two classes. Experimental results in Sec.~\ref{sec:adv_exps} will show the empirical effectiveness of introducing the extra generator.

\section{Experiments}\label{sec:exp}
In this section, we first demonstrate the superiority of our method on the task of deepfake detection through a large number of experiments, and then we illustrate the effectiveness of our Gaussian blurring adversarial attack through white box attack experiments on ImageNet~\cite{imagenet}.

\begin{table*}[htb]
    \setlength\tabcolsep{5.0pt} 
        \footnotesize
      \caption{Comparison between different adversarial training settings in improving the generalization to unseen forgery technologies. All models were trained on the NT C23 data in FF++~\cite{rossler2019faceforensics++} and tested on data generated using different technologies (indicated by ``NT $\rightarrow\ast$''). DFD is extremely imbalanced so we only report the AUC scores on it, since the overall accuracy makes less sense on imbalanced data. }\vskip -0.1in
      \label{table:ablation type}
      \begin{center}
        \begin{tabular}{c|c|c|c|c|c|c|c|c|c|c|c|c|c|c}
            \hline 
            Method & \multicolumn{12}{c}{NT $\rightarrow\ast$} \\
            \hline
            & \multicolumn{2}{c|}{NT} & \multicolumn{2}{c|}{DF} & \multicolumn{2}{c|}{F2F} & \multicolumn{2}{c|}{FS} & \multicolumn{2}{c|}{DFD} & \multicolumn{2}{c|}{Celeb-DF} & \multicolumn{2}{c}{Avg} \\
            \hline
            & AUC & ACC & AUC & ACC & AUC & ACC & AUC & ACC & AUC & ACC & AUC & ACC & AUC & ACC \\
            & (\%) & (\%) & (\%) & (\%) & (\%) & (\%) & (\%) & (\%) & (\%) & (\%) & (\%) & (\%) & (\%) & (\%) \\
            \hline 
            EfficientNet~\cite{efficientnet} & \textbf{98.75} & \textbf{95.40} & 83.75 & 60.42 & 61.15 & 51.64 & 43.58 & 48.74 & 74.34 &  --  & 57.48 & 57.32 & 69.84 & 62.70\\
            \hline 
            + Grad-AAT & 98.00 & 93.69 & 85.91 & 59.13 & 71.12 & 52.54 & 44.97 & 49.33 & 75.38 &  --  & 62.17 & 61.00 & 72.93 & 63.14\\
            \hline 
            + Grad-SAT & 97.83 & 93.40 & 85.03 & 59.13 & 67.44 & 51.99 & 44.60 & 49.01 & 75.19 &  --  & 60.05 & 59.07 & 71.69 & 62.52\\
            \hline 
            + Grad-BAT & 98.38 & 94.60 & 87.09 & 60.62 & 72.50 & 53.91 & 47.00 & 50.21 & 77.02 &  --  & 63.13 & 65.05 & 74.19 & 64.88\\
            \hline 
            + Gen-BAT & 98.12 & 94.96 & 87.50 & 67.52 & 69.82 & 54.66 & 47.12 & 50.04 & 77.02 &  --  & 66.05 & 66.90 & 74.27 & 66.82\\
            \hline 
            + Two-Gen-BAT & 98.72 & 95.26 & 87.51 & 69.40 & 74.65 & 56.19 & 48.99 & 50.43 & 76.60 &  --  & 66.84 & 67.91 & 75.55 & 67.84\\
            \hline 
            \hline
            + Combined AT & 98.40 & 94.95 & \textbf{88.90} & \textbf{71.08} & \textbf{76.13} & \textbf{57.90} & \textbf{50.13} & \textbf{51.14} & \textbf{77.74} & -- & \textbf{68.45} & \textbf{69.00} & \textbf{76.63} & \textbf{68.81}\\
            \hline

        \end{tabular}
    \end{center}
\end{table*}

\begin{table*}[htb]
    \setlength\tabcolsep{2.0pt} 
        \footnotesize
      \caption{Comparison between different adversarial training settings in improving the generalization to unseen image/video qualities (indicated by $\rightarrow$). The train and test data were split from the NT data in FF++~\cite{rossler2019faceforensics++}, of possibly different qualities. The $\rightarrow$ symbol indicates the models were trained on a specific image/video quality, and were expected to generalize to other image/video qualities.} \vskip -0.15in
      \label{table:ablation quality}
      \begin{center}
        \begin{tabular}{c|c|c|c|c|c|c|c|c|c|c|c|c|c|c|c|c|c|c}
            \hline
            Method & \multicolumn{6}{c|}{Raw $\rightarrow\ast$} & \multicolumn{6}{c|}{C23 $\rightarrow\ast$} & \multicolumn{6}{c}{C40 $\rightarrow\ast$} \\
            \hline
             & \multicolumn{2}{c|}{Raw} & \multicolumn{2}{c|}{C23} & \multicolumn{2}{c|}{C40} & \multicolumn{2}{c|}{Raw} & \multicolumn{2}{c|}{C23} & \multicolumn{2}{c|}{C40} & \multicolumn{2}{c|}{Raw} & \multicolumn{2}{c|}{C23} & \multicolumn{2}{c}{C40}\\
            \hline
            & AUC & ACC & AUC & ACC & AUC & ACC & AUC & ACC & AUC & ACC & AUC & ACC & AUC & ACC & AUC & ACC & AUC & ACC \\
            & (\%) & (\%) & (\%) & (\%) & (\%) & (\%) & (\%) & (\%) & (\%) & (\%) & (\%) & (\%) & (\%) & (\%) & (\%) & (\%) & (\%) & (\%) \\
            \hline 
            EfficientNet~\cite{efficientnet} & \textbf{99.51}  & \textbf{99.24} & 66.35 & 51.05 & 56.03 & 50.60 & 98.41 & 94.09 & \textbf{98.75}  & \textbf{95.40} & 69.40 & 54.85 & 86.27 & 78.36 & 90.28 & 80.85 & 89.95  & 81.65 \\ 
            \hline
            + Grad-AAT & 99.11 & 99.04 & 66.49 & 53.02 & 56.42 & 50.83 & 97.34 & 93.12 & 98.00 & 93.69 & 69.01 & 53.93 & 86.33 & 78.42 & 87.35 & 79.00 & 88.69 & 80.15 \\ 
            \hline
            + Grad-SAT & 99.02 & 98.99 & 65.44 & 50.70 & 55.33 & 50.45 & 97.21 & 92.80 & 97.83 & 93.40 & 67.46 & 51.33 & 86.00 & 76.49 & 86.66 & 78.44 & 87.84 & 79.93 \\ 
            \hline
            + Grad-BAT & 99.05 & 99.10 & 92.90 & 68.84 & 60.51 & 54.62 & 98.80 & 94.80 & 98.38 & 94.60 & 73.55 & 61.13 & 89.75 & 79.70 & 90.67 & 81.95 & 89.05 & 80.52 \\ 
            \hline
            + Gen-BAT & 98.80 & 97.75 & 95.52 & 80.71 & 68.53 & 56.71 & 98.77 & 94.95 & 98.12 & 94.96 & 73.71 & 61.84 & 89.81 & 80.93 & 91.54 & 83.48 & 89.00 & 81.05 \\ 
            \hline
            + Two-Gen-BAT & 99.46 & 99.04 & \textbf{95.93} & \textbf{84.19} & \textbf{72.18} & \textbf{61.87} & \textbf{98.92} & \textbf{95.52} & 98.72 & 95.26 & 74.73 & 62.53 & \textbf{90.58} & \textbf{82.91} & \textbf{93.51} & \textbf{85.36} & \textbf{94.19} & \textbf{87.02} \\ 
            \hline
            \hline
            + Combined AT & 99.41 & 98.95 & 95.80 & 82.99 & 71.02 & 59.82 & 98.73 & 94.80 & 98.40 & 94.95 & \textbf{75.43} & \textbf{62.93} & 89.99 & 82.00 & 92.80 & 84.95 & 92.00 & 85.84 \\ 
            \hline
        \end{tabular}
    \end{center}
\end{table*}

\textbf{Dataset.}
FaceForensics++ (FF++)~\cite{rossler2019faceforensics++} is a recently released large scale deepfake video detection dataset, containing 1,000 real videos, in which 720 videos were used for training, 140 videos were reserved for verification, and 140 videos were used for test. Each real video in the dataset was manipulated using four advanced methods, including DeepFakes (DF)~\cite{deepfake}, Face2Face (F2F)~\cite{face2face}, FaceSwap (FS)~\cite{faceswap_project}, and NeuralTextures (NT)~\cite{neuraltexture}, to generate four fake videos. 
We followed the official split of the training, validation, and test sets in our experiments.
Each video in the dataset were processed to have three video qualities, namely RAW, C23 (which is compressed from the raw data but has relatively high quality), and C40 (which is compressed to have relatively low quality). 
For each quality, there are 5,000 (real and fake) videos in total, and we extracted 270 frames from each video following the official implementation of face detection and alignment in~\cite{rossler2019faceforensics++}.
In order to evaluate the generalization ability of the model, we trained models on videos from one specific method and evaluated on those generated by a variety of manipulation methods and image/video qualities. 

To make the evaluation more comprehensive, we introduced two more deepfake datasets: DFD~\cite{dfd} and Celeb-DF~\cite{li2020celeb}. DFD contains 3,068 deepfake videos, which were forged based on 363 real videos. We used all the real and fake videos, and we randomly selected 10 frames from each of them for testing. Celeb-DF contains 590 real videos and 5,639 fake videos, and we used its official test set.

\textbf{Implementation Details.}
We applied our adversarial training to existing deepfake detection models to testify its effectiveness.
We first considered EfficientNet~\cite{efficientnet} (which is a common choice of many winning solutions to the Deepfake Detection Challenge~\footnote{\url{https://www.kaggle.com/c/deepfake-detection-challenge}}).
EfficientNet was originally designed for image classification and it was transferred to the deepfake detection task by replacing its highest fully-connected layer with one that outputs two dimensional logits.
This layer was randomly initialized and the other layers of the model were all pre-trained on ImageNet~\cite{imagenet}. We used the RAdam~\cite{radam} optimizer with $\beta_1=0.9$ and $\beta_2=0.999$ to train the model. We adopted a weight decay of $5 \times 10^{-4}$. The learning rate was initialized to $5 \times 10^{-4}$ and decayed by $0.1$ every $10$ epochs. For models trained along with generators, the learning rate of the generators was initialized to $2 \times 10^{-3}$. For experiments involving Gaussian blur, we set the blur kernel size $k$ to $9$, unless otherwise clarified. To ensure numerical stability, we optimize or generate $1/\sigma^{adv}$ rather than $\sigma^{adv}$ for BAT in practice. All experiments were performed in a PyTorch~\cite{pytorch} environment, running with an Intel Xeon Gold 6130 CPU and an Nvidia Tesla V100 GPU. 
Besides EfficientNet, we also considered an Xception~\cite{xception} model following~\cite{rossler2019faceforensics++} and two recent models proposed by Stehouwer \etal~\cite{stehouwer2019detection} and Zhao \etal~\cite{zhao2021multiattentional} respectively
We used the prediction accuracy and the area under the receiver operating characteristic curve (AUC) as evaluation metrics. 
Following prior arts, we report the AUC scores in percentage for comparison in the paper.

\begin{table*}[tb]
    \setlength\tabcolsep{0.5pt} 
        \footnotesize
      \caption{When training was perform on NT data of all qualities (\ie, RAW, C23, and C40), and test was performed on the same qualities.}
      \vskip -0.18in
      \label{table:ablation quality mix}
      \begin{center}
        \begin{tabular}{p{2cm}<{\centering}|p{2cm}<{\centering}|p{2cm}<{\centering}|p{2cm}<{\centering}|p{2cm}<{\centering}|p{2cm}<{\centering}|p{2cm}<{\centering}|p{2cm}<{\centering}}
            \hline
            Metric & EfficientNet~\cite{efficientnet} & + Grad-AAT & + Grad-SAT & + Grad-BAT & + Gen-BAT & + Two-Gen-BAT & + Combined AT\\
            \hline
            AUC (\%) & 88.81 & 88.08 & 87.47 & 89.95 & 90.20 & 90.36 & \textbf{90.45} \\
            \hline 
            ACC (\%) & 81.35 & 80.20 & 80.11 & 83.25 & 83.67 & 84.10 & \textbf{85.02} \\
            \hline
            
        \end{tabular} \vskip -0.5in
    \end{center}
\end{table*}

\begin{table*}[htbp]
    \setlength\tabcolsep{5.0pt} 
        \footnotesize
      \caption{Comparison between adversarial training and traditional data augmentation in improving generalization to unseen forgery technologies. The models were trained on the NT C23 data and tested on data generated using various technologies (indicated by ``NT $\rightarrow\ast$'').}
      \vskip -0.15in
      \label{table:aug forgery type}
      \begin{center}
        \begin{tabular}{c|c|c|c|c|c|c|c|c|c|c|c|c|c|c}
            \hline 
            Method & \multicolumn{12}{c}{NT $\rightarrow\ast$} \\
            \hline
            & \multicolumn{2}{c|}{NT} & \multicolumn{2}{c|}{DF} & \multicolumn{2}{c|}{F2F} & \multicolumn{2}{c|}{FS} & \multicolumn{2}{c|}{DFD} & \multicolumn{2}{c|}{Celeb-DF} & \multicolumn{2}{c}{Avg} \\
            \hline
            & AUC & ACC & AUC & ACC & AUC & ACC & AUC & ACC & AUC & ACC & AUC & ACC & AUC & ACC \\
            & (\%) & (\%) & (\%) & (\%) & (\%) & (\%) & (\%) & (\%) & (\%) & (\%) & (\%) & (\%) & (\%) & (\%) \\
            \hline 
            EfficientNet~\cite{efficientnet} & 98.75 & 95.40 & 83.75 & 60.42 & 61.15 & 51.64 & 43.58 & 48.74 & 74.34 &  --  & 57.48 & 57.32 & 69.84 & 62.70 \\
            \hline
            + Gaussian Noise & 98.66 & 95.19 & 83.78 & 60.49 & 61.19 & 51.76 & 43.68 & 48.85 & 74.44 & -- &  58.06 & 57.96 & 69.97 & 62.85\\
            \hline 
            + Gaussian Blur & 98.78 & 95.95 & 83.84 & 60.48 & 61.17 & 51.81 & 43.64 & 48.86 & 74.45 &  --  & 58.43 & 58.24 & 70.05 & 63.07\\
            \hline 
            + JPEG Compression & 98.64 & 95.14 & 83.78 & 60.45 & 61.17 & 51.76 & 43.62 & 48.82 & 74.45 &  --  & 58.19 & 58.01 & 69.98 & 62.84\\
            \hline 
            + Combined Traditional & \textbf{98.85} & \textbf{95.98} & 83.86 & 60.48 & 61.26 & 51.83 & 43.67 & 48.89 & 74.56 & -- & 58.55 & 58.49 & 70.13 & 63.13\\
            \hline
            + Two-Gen-BAT (ours) & 98.72 & 95.26 & 87.51 & 69.40 & 74.65 & 56.19 & 48.99 & 50.43 & 76.60 &  --  & 66.84 & 67.91 & 75.55 & 67.84\\
            \hline 
            + Combined AT  (ours) & 98.40 & 94.95 & \textbf{88.90} & \textbf{71.08} & \textbf{76.13} & \textbf{57.90} & \textbf{50.13} & \textbf{51.14} & \textbf{77.74} & -- & \textbf{68.45} & \textbf{69.00} & \textbf{76.63} & \textbf{68.81}\\
            \hline

        \end{tabular}
    \end{center}
\end{table*}

\begin{table*}[!htbp]
    \setlength\tabcolsep{2.0pt} 
        \footnotesize
      \caption{Comparison between adversarial training and the traditional data augmentation in improving generalization to unseen image/video qualities. The train and test data were split from the NT data in FF++~\cite{rossler2019faceforensics++}, with possibly different qualities. The $\rightarrow$ symbol indicates the models were trained on a specific image/video quality, and were expected to generalize to other image/video qualities.}\vskip -0.15in
      \label{table:aug image quality}
      \begin{center}
        \begin{tabular}{c|c|c|c|c|c|c|c|c|c|c|c|c|c|c|c|c|c|c}
            \hline
            Method & \multicolumn{6}{c|}{Raw $\rightarrow\ast$} & \multicolumn{6}{c|}{C23 $\rightarrow\ast$} & \multicolumn{6}{c}{C40 $\rightarrow\ast$} \\
            \hline
             & \multicolumn{2}{c|}{Raw} & \multicolumn{2}{c|}{C23} & \multicolumn{2}{c|}{C40} & \multicolumn{2}{c|}{Raw} & \multicolumn{2}{c|}{C23} & \multicolumn{2}{c|}{C40} & \multicolumn{2}{c|}{Raw} & \multicolumn{2}{c|}{C23} & \multicolumn{2}{c}{C40}\\
            \hline
            & AUC & ACC & AUC & ACC & AUC & ACC & AUC & ACC & AUC & ACC & AUC & ACC & AUC & ACC & AUC & ACC & AUC & ACC \\
            & (\%) & (\%) & (\%) & (\%) & (\%) & (\%) & (\%) & (\%) & (\%) & (\%) & (\%) & (\%) & (\%) & (\%) & (\%) & (\%) & (\%) & (\%) \\
            \hline 
            EfficientNet~\cite{efficientnet} & \textbf{99.51} & 99.24 & 66.35 & 51.05 & 56.03 & 50.60 & 98.41 & 94.09 & 98.75  & 95.40 & 69.40 & 54.85 & 86.27 & 78.36 & 90.28 & 80.85 & 89.95  & 81.65 \\ 
            \hline
            + Gaussian Noise & 99.38 & 99.17 & 69.66 & 54.75 & 57.24 & 50.97 & 98.47 & 94.25 & 98.66 & 95.19 & 70.04 & 56.18 & 87.50 & 78.98 & 90.78 & 81.13 &89.78 & 80.55\\ 
            \hline
            + Gaussian Blur & 99.47 & 99.24 & 77.63 & 57.46 & 59.24 & 51.29 & 98.88 & 94.65 & 98.78 & 95.95 & 71.08 & 57.89 & 89.54 & 79.83 & 90.92 & 82.03 & 90.75 & 81.83 \\ 
            \hline
            + JPEG Compression & 99.39 & 99.17 & 95.07 & 80.19 & 67.95 & 56.18 & 98.25 & 94.30 & 98.64 & 95.14 & 73.19 & 61.21 & 89.33 & 80.46 & 91.19 & 83.20 & 91.95 & 82.66 \\ 
            \hline
            + Combined Traditional & 99.46 & \textbf{99.26} & 83.57 & 73.52 & 68.11 & 57.29 & \textbf{98.98} & 94.90 & \textbf{98.85} & \textbf{95.98} & 73.26 & 61.58 & 89.69 & 80.78 & 91.75 & 83.36 & 92.07 & 83.75 \\ 
            \hline 
            + Two-Gen-BAT (ours) & 99.46 & 99.04 & \textbf{95.93} & \textbf{84.19} & \textbf{72.18} & \textbf{61.87} & 98.92 & \textbf{95.52} & 98.72 & 95.26 & 74.73 & 62.53 & \textbf{90.58} & \textbf{82.91} & \textbf{93.51} & \textbf{85.36} & \textbf{94.19} & \textbf{87.02} \\ 
            \hline
            + Combined AT (ours) & 99.41 & 98.95 & 95.80 & 82.99 & 71.02 & 59.82 & 98.73 & 94.80 & 98.40 & 94.95 & \textbf{75.43} & \textbf{62.93} & 89.99 & 82.00 & 92.80 & 84.95 & 92.00 & 85.84 \\ 
            \hline
            
        \end{tabular}
    \end{center}
\end{table*}

\subsection{Different Settings for Adversarial Training}\label{sec:adv_exps}

\begin{figure}
    \begin{center}
        \includegraphics[scale=0.135]{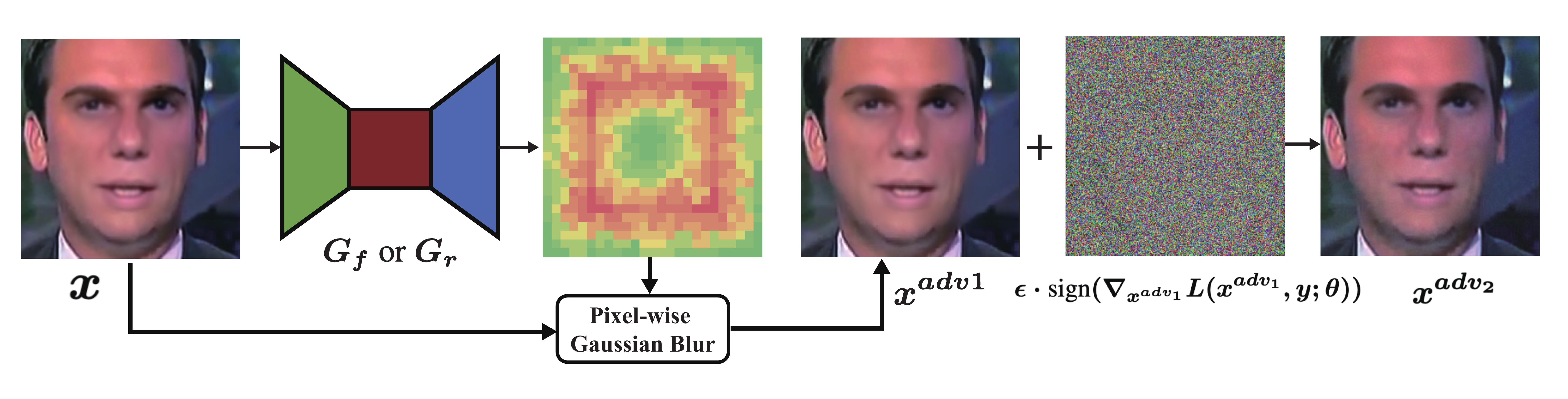}
    \end{center}
    \vskip -0.15in
    \caption{Our combined adversarial training. First, we generate the $\sigma^{adv}$ map for performing adversarial blurring, then we craft $x^{adv_1}$ following the right panel of Figure~\ref{fig:overview} and finally incorporate the additive perturbation to get $x^{adv_2}$ from $x^{adv_1}$.}\vskip -0.15in
    \label{fig:combine}
\end{figure}

Since several different ways of generating adversarial examples and performing adversarial training have been introduced, we compare them first, including: (\romannumeral 1)  Grad-AAT: input-gradient-based additive adversarial training, (\romannumeral 2) Grad-SAT: input-gradient-based spatially-transformed adversarial training, (\romannumeral 3) Grad-BAT: input-gradient-based blurring adversarial training, (\romannumeral 4) Gen-BAT: generator-based blurring adversarial training, (\romannumeral 5) Two-Gen-BAT: BAT with two generators, and, partially inspired by a general image degeneration process, (\romannumeral 6) Combined AT: combining Two-Gen-BAT with Grad-AAT (see Figure~\ref{fig:combine} for an overview). All competing models were trained on the NT data only and might be tested on the other fake data. Note that generator-based AAT and generator-based SAT did not show any improvement to the baseline solution in our experiment thus their results will not be shown. The number of real and fake videos in DFD~\cite{dfd} is seriously unbalanced, so we did not show the accuracy on DFD dataset. 

It can be seen from Table~\ref{table:ablation type} that the considered various types of adversarial training all improve the generalization ability of the obtained models to unseen forgery types (\ie, DF, F2F, FS, DFD, and Celeb-DF). Table~\ref{table:ablation quality} further shows that equipped with the operation of pixel-wise blurring, our Grad-BAT, Gen-BAT, and the combined AT also improve the generalization performance of the obtained models for different image/video qualities, though other adversarial training methods can hardly contribute under such circumstance. 
Also, the results demonstrate the performance of our BAT equipped with two generators is superior to that only equipped with a single generator. Moreover, although adversarial training leads to slightly degraded performance when the training and test data come from the same deepfake forgery technology and share the same image/video quality (\cf the first column of Tables~\ref{table:ablation type} and~\ref{table:ablation quality}), introducing two generators mitigate such performance degradation to some extent.
We also considered when data of different image/video qualities was trained altogether, and test was performed on all the qualities. Somewhat surprisingly, we found that the performance degradation (observed in Table~\ref{table:ablation quality}) was well-mitigated with our adversarial training (see Table~\ref{table:ablation quality mix}).

The experiment in this section is mostly performed based on EfficientNet~\cite{efficientnet}, a common choice of many winning deepfake detection solutions. Similar observations can be made on other deepfake detection models as well, \eg, Xception~\cite{xception}. See Appendix~\ref{sec:sup xception} for detailed results on Xception. In addition to the single-frame deepfake detection models, we also tested with multi-frame models using TSN~\cite{wang2018temporal}, and similar conclusions can be obtained. Specifically, when trained using C40 data, the classification AUC of the TSN baseline is $89.24\%$, $92.21\%$, and $91.34\%$ on Raw, C23, and C40 data, respectively, while with our method, the results are $92.03\%$, $94.49\%$, and $93.58\%$. Generalization to unseen forgery technologies is also improved in the multi-frame setting, from $72.39\%$ to $76.80\%$ on average AUC.

\subsection{Comparison to Other Methods}


\textbf{Adversarial training vs data augmentation.}
Adversarial training can be regarded as an advanced way of performing data augmentation. On this point, we further compare the proposed method to some traditional data augmentation strategies, including  Gaussian noise, traditional Gaussian blurring, and JPEG compression~\cite{wang2020cnn}. Tables~\ref{table:aug forgery type} and~\ref{table:aug image quality} compare adversarial training (\ie, Two-Gen-BAT and combined AT) to these strategies and their combination (named ``combined Traditional'' in the tables). It can be seen that these traditional data augmentation strategies and their combination hardly improve model generalization to unseen forgery technologies, despite some unsurprising improvement in generalization to unseen image/video qualities. In addition to the results on EfficientNet, see Tables \ref{table:sup aug type} and \ref{table:sup aug quality} in the appendices for similar results based on Xception. 

\textbf{Incorporating with other deepfake detection models.}
We would also like to see whether our adversarial training could improve more advanced models (than Xception~\cite{xception} and EfficientNet~\cite{efficientnet}) similarly. 
We tested with the ones proposed by Stehouwer \etal~\cite{stehouwer2019detection} and Zhao \etal~\cite{zhao2021multiattentional} respectively, and we tried applying our combined AT and Two-Gen-BAT on them. The whole FF++ training set (including DF, F2F, FS, and NT data of all qualities) was used for training to better suit the model setting in~\cite{stehouwer2019detection}, and we tested the model performance on DFD, Celeb-DF, and the test set of FF++. It can be seen from Table~\ref{table:other method} that the adversarially trained models generalize considerably better on unseen forgery technologies, \ie, DFD and Celeb-DF.

\begin{table}[!htbp]
    \setlength\tabcolsep{3.0pt} 
      \footnotesize
      \caption{Evaluation of our adversarial training on different recent deepfake detection baseline models. All these models were trained on the FF++ data of all qualities (\ie, RAW, C23, and C40)~\cite{rossler2019faceforensics++} and tested on FF++~\cite{rossler2019faceforensics++}, DFD~\cite{dfd}, and Celeb-DF~\cite{li2020celeb}.}\vskip -0.15in
      \label{table:other method}
      \begin{center}
        \begin{tabular}{c|c|c|c|c|c|c}
            \hline 
            Method & \multicolumn{6}{c}{FF++ $\rightarrow\ast$} \\
            \hline
            & \multicolumn{2}{c|}{FF++} & \multicolumn{2}{c|}{DFD} & \multicolumn{2}{c}{Celeb-DF}\\
            \hline
            & AUC & ACC & AUC & ACC & AUC & ACC\\
            & (\%) & (\%) & (\%) & (\%) & (\%) & (\%) \\
            \hline 
            Xception~\cite{xception} & 96.77 & 92.51 & 84.85 & - & 55.11 & 54.88\\
            \hline
            + Two-Gen-BAT (ours) & 97.45 & 93.25 & 86.49 & - & 69.28 & 63.17\\
            \hline
            + Combined AT (ours) & \textbf{97.62} & \textbf{93.38} & \textbf{87.10} & - & \textbf{70.61} & \textbf{64.85}\\
            \hline
            \hline
            EfficientNet~\cite{efficientnet} & 97.58 & 93.49 & 85.67 & - & 56.87 & 55.89\\
            \hline
            + Two-Gen-BAT (ours) & 98.46 & 94.00 & 87.58 & - & 72.69 & 66.46\\
            \hline
            + Combined AT (ours) & \textbf{98.62} & \textbf{94.12} & \textbf{87.95} & - & \textbf{73.46} & \textbf{67.52}\\
            \hline
            \hline
            Stehouwer \etal~\cite{stehouwer2019detection} & 96.79 & 94.17 & 86.84 & - & 59.74 & 59.36\\
            \hline
            + Two-Gen-BAT (ours) & 97.57 & \textbf{95.16} & 88.84 & - & 74.56 & 69.50\\
            \hline
            + Combined AT (ours) & \textbf{97.59} & 95.11 & \textbf{89.33} & - & \textbf{76.03} & \textbf{70.45}\\
            \hline
            \hline
            Zhao \etal~\cite{zhao2021multiattentional} & 96.84 & 94.46 & 87.46 & - & 60.92 & 60.27\\
            \hline
            + Two-Gen-BAT (ours) & 97.59 & \textbf{95.85} & 89.10 & - & 75.61 & 70.00\\
            \hline
            + Combined AT (ours) & \textbf{98.00} & \textbf{95.85} & \textbf{89.78} & - & \textbf{76.82} & \textbf{70.54}\\
            \hline
        \end{tabular}
    \end{center}
\end{table}

\subsection{Adversarial Blurring as An Attack}

We also tested how adversarial blurring performed as an attack. 
Following prior work, we used the ImageNet~\cite{imagenet} data and models to test the attack. We selected 50,000 images from the official ImageNet validation set and crafted adversarial examples on the basis of these benign images. 
We adopted the adversarial accuracy (\ie, the prediction accuracy of the victim model on adversarial examples) as an evaluation metric, 
and we chose three ImageNet models for the experiment, namely Inception v3 (Inc-v3)~\cite{szegedy2016rethinking} (top-1 accuracy: 77.21\%), Inception v4 (Inc-v4)~\cite{szegedy2016inception} (top-1 accuracy: 80.12\%), and Inception-ResNet v2 (IncRes-v2)~\cite{szegedy2016inception} (top-1 accuracy: 80.33\%). We tested our (single-step) adversarial blurring and FGSM~\cite{goodfellow2014explaining} in this experiment\footnote{To save space, we provide their iterative results in Appendix~\ref{sec:sup kernel-size}.}, and we specifically tested the prediction accuracy of a model on adversarial examples crafted using other models, as an evaluation of the adversarial transferability (or say generalization). Table~\ref{table:attack} summarizes these results, and it can be seen that the adversarially blurred examples transfer reasonably well across ImageNet models.
We let the kernel size be $5$ and $\epsilon$ be 0.1 for our adversarial blurring, while for FGSM, we let $\epsilon=16/255$. All entries of $\sigma$ in Eq.~(\ref{equ:one_step}) were set to $1$.
More results can be found in the Appendix~\ref{sec:sup attack}. 

\begin{table}[!htbp]
    \setlength\tabcolsep{3.0pt} 
        \footnotesize
      \caption{Adversarial accuracy of ImageNet models on our adversarial examples and the FGSM examples.}\vskip -0.15in
      \label{table:attack}
      \begin{center}
        \begin{tabular}{c|c|c|c|c|c|c}
            \hline 
            Source & \multicolumn{6}{c}{Target} \\
             \hline
             & \multicolumn{2}{c|}{Inc-v3} & \multicolumn{2}{c|}{Inc-v4} & \multicolumn{2}{c}{IncRes-v2}\\
            \hline
             & FGSM & BAT & FGSM & BAT & FGSM & BAT\\
            \hline
            Inc-v3 & 25.02\% & 21.48\% & 72.46\% & 71.04\% & 72.74\% & 71.18\% \\
            \hline 
            Inc-v4 & 69.36\% & 65.86\% & 36.92\% & 30.84\% & 72.36\% & 71.18\% \\
            \hline
            IncRes-v2 & 69.60\% & 65.84\% & 71.62\% & 71.04\% & 44.38\% & 33.20\%\\
            \hline
        \end{tabular}
    \end{center}
\end{table}

\section{Conclusion}
We aim at improving the generalization ability of deepfake detection models, by introducing adversarial training to them. Towards the goal, we have developed a new type of adversarial attacks on the basis of image blurring, which can be effectively and efficiently crafted by introducing two generators for training the deepfake detection models. Our method encourages models to learn essential and generalizable features to distinguish fake from real, rather than those obvious artifacts that are too specific. The proposed adversarial blurring-based method can further be combined with the other adversarial methods (\eg, FGSM) to achieve further improvement in generalization. Extensive experiments have shown that our adversarial training improves the generalization ability of deepfake detection models to unseen image/video qualities and unseen deepfake technologies. The performance of the proposed adversarial examples in attacking ImageNet models has also been tested.


%


%

\appendices


\section{Using Xception As A Baseline}\label{sec:sup xception}
Here we report the results of models that use Xception~\cite{xception} as backbone.

Table~\ref{table:sup type} and Table~\ref{table:sup quality} compare a variety of plausible settings of adversarial training. Consistent with the results in our main paper, here we still compare: (\romannumeral 1)  Grad-AAT: input-gradient-based additive adversarial training, (\romannumeral 2) Grad-SAT: input-gradient-based spatially-transformed adversarial training, (\romannumeral 3) Grad-BAT: input-gradient-based blurring adversarial training, (\romannumeral 4) Gen-BAT: generator-based blurring adversarial training, (\romannumeral 5) Two-Gen-BAT: our BAT with two generators, and, partially inspired by a general image degeneration process, (\romannumeral 6) Combined AT: combining Two-Gen-BAT with Grad-AAT. It can be seen that, similar to the observations on EfficientNet~\cite{efficientnet}, our BAT with two generators (\ie, Two-Gen-BAT) achieves significantly superior performance in comparison with many other settings, and, when combined with Grad-AAT, further improvement can be obtained.

Table~\ref{table:sup aug type} and Table~\ref{table:sup aug quality} compare our adversarial training (\ie, Two-Gen-BAT and combined AT) to the traditional data augmentation strategies together with their combinations, \ie, the traditional Gaussian noise, traditional Gaussian blur, JPEG compression~\cite{wang2020cnn}, and ``combined traditional'').

\section{More Results for Attack}\label{sec:sup attack}
In addition to the results reported in Section~\ref{sec:exp}, we further report results for performing attacks with different settings of the blurring kernel. We also provide iterative version of BAT and FGSM (or say PGD) for attack (see Table \ref{table:iter attack}). 

\begin{figure}[!htbp]
    \begin{center}
        \includegraphics[scale=0.145]{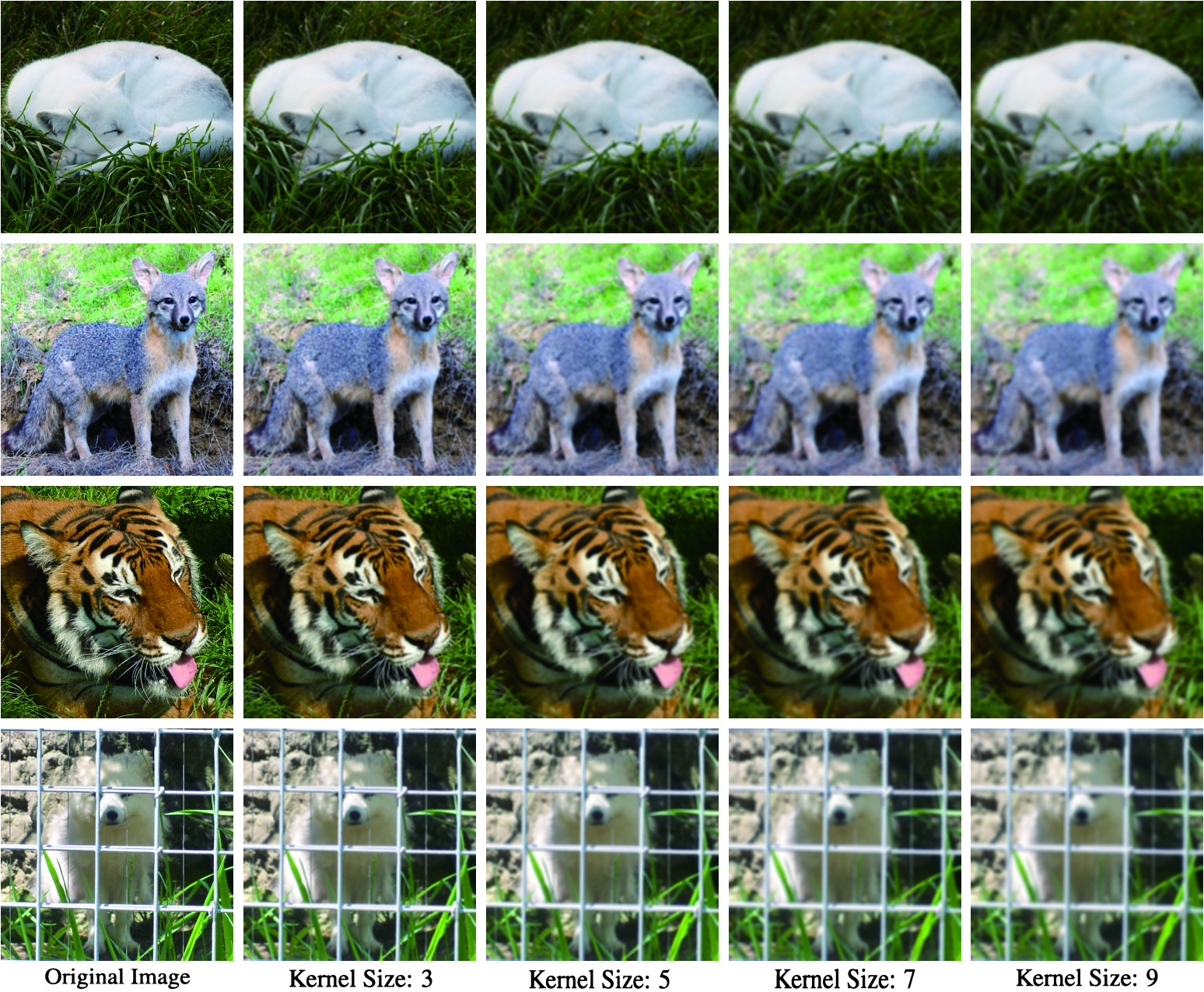}
    \end{center}
    \vskip -0.15in
    \caption{Adversarial blurred examples crafted with different kernel sizes. Obviously, the image becomes more and more blurry as the kernel size increases. Zoom in for better observation.} 
    \label{fig: kernel}
\end{figure}

\begin{table*}[!htbp]
    \setlength\tabcolsep{5.0pt} 
        \footnotesize
      \caption{Comparison between different adversarial training settings in improving generalization to unseen face forgery technologies (based on Xception~\cite{xception}). All models were trained on the NT C23 data in FF++~\cite{rossler2019faceforensics++} and tested on data generated using different forgery technologies (indicated by ``NT $\rightarrow\ast$''). The DFD dataset is extremely imbalanced so we only report the AUC scores on it, since the overall accuracy makes less sense on imbalanced data. } \vskip -0.15in
      \label{table:sup type} 
      \begin{center}
        \begin{tabular}{c|c|c|c|c|c|c|c|c|c|c|c|c|c|c}
            \hline 
            Method & \multicolumn{12}{c}{NT $\rightarrow\ast$} \\
            \hline
            & \multicolumn{2}{c|}{NT} & \multicolumn{2}{c|}{DF} & \multicolumn{2}{c|}{F2F} & \multicolumn{2}{c|}{FS} & \multicolumn{2}{c|}{DFD} & \multicolumn{2}{c|}{Celeb-DF} & \multicolumn{2}{c}{Avg} \\
            \hline
            & AUC & ACC & AUC & ACC & AUC & ACC & AUC & ACC & AUC & ACC & AUC & ACC & AUC & ACC \\
            & (\%) & (\%) & (\%) & (\%) & (\%) & (\%) & (\%) & (\%) & (\%) & (\%) & (\%) & (\%) & (\%) & (\%) \\
            \hline 
            Xception~\cite{xception} & 98.06 & \textbf{94.26} & 82.66 & 60.54 & 60.96 & 51.07 & 42.94 & 48.37 & 69.82 &  --  & 53.90 & 53.75 & 68.06 & 61.60\\
            \hline 
            + Grad-AAT & 97.71 & 93.42 & 84.10 & 60.91 & 65.68 & 51.85 & 43.55 & 48.72 & 71.09 &  --  & 56.88 & 55.98 & 69.84 & 62.18\\
            \hline 
            + Grad-SAT & 97.56 & 93.33 & 82.89 & 60.53 & 63.21 & 51.43 & 43.06 & 48.40 & 70.34 &  --  & 56.29 & 54.70 & 68.89 & 61.68\\
            \hline 
            + Grad-BAT & 98.40 & 93.80 & 85.57 & 62.10 & 68.45 & 52.96 & 44.61 & 49.93 & 72.15 &  --  & 58.44 & 57.42 & 71.27 & 63.57\\
            \hline 
            + Gen-BAT & 98.44 & 93.91 & 86.09 & 64.13 & 69.42 & 53.81 & 45.10 & 49.94 & 73.10 &  --  & 59.33 & 59.05 & 71.91 & 64.17\\
            \hline 
            + Two-Gen-BAT & 98.61 & 94.15 & 86.18 & 64.54 & 72.02 & 55.35 & 45.87 & 50.47 & 73.15 &  --  & 59.69 & 60.29 & 72.59 & 65.16\\
            \hline 
            \hline
            + Combined AT & \textbf{98.68} & 93.99 & \textbf{86.58} & \textbf{65.42} & \textbf{72.75} & \textbf{55.58} & \textbf{46.16} & \textbf{50.80} & \textbf{74.45} & -- & \textbf{61.47} & \textbf{62.06} & \textbf{73.35} & \textbf{65.57}\\
            \hline
        \end{tabular}
    \end{center}
\end{table*}

\begin{table*}[!htbp]
    \setlength\tabcolsep{2.0pt} 
        \footnotesize
      \caption{Comparison between different adversarial training settings for improving generalization to unseen image/video qualities (based on Xception~\cite{xception}). All models were trained on NT data in FF++~\cite{rossler2019faceforensics++}, and tested on NT data of possibly different qualities. The $\rightarrow$ symbol indicates the models were trained on a specific image/video quality, and were expected to generalize to other image/video qualities.}
      \vskip -0.15in
      \label{table:sup quality}
      \begin{center}
        \begin{tabular}{c|c|c|c|c|c|c|c|c|c|c|c|c|c|c|c|c|c|c}
            \hline
            Method & \multicolumn{6}{c|}{Raw $\rightarrow\ast$} & \multicolumn{6}{c|}{C23 $\rightarrow\ast$} & \multicolumn{6}{c}{C40 $\rightarrow\ast$} \\
            \hline
             & \multicolumn{2}{c|}{Raw} & \multicolumn{2}{c|}{C23} & \multicolumn{2}{c|}{C40} & \multicolumn{2}{c|}{Raw} & \multicolumn{2}{c|}{C23} & \multicolumn{2}{c|}{C40} & \multicolumn{2}{c|}{Raw} & \multicolumn{2}{c|}{C23} & \multicolumn{2}{c}{C40}\\
            \hline
            & AUC & ACC & AUC & ACC & AUC & ACC & AUC & ACC & AUC & ACC & AUC & ACC & AUC & ACC & AUC & ACC & AUC & ACC \\
            & (\%) & (\%) & (\%) & (\%) & (\%) & (\%) & (\%) & (\%) & (\%) & (\%) & (\%) & (\%) & (\%) & (\%) & (\%) & (\%) & (\%) & (\%) \\
            \hline 
            Xception~\cite{xception} & \textbf{99.21}  & \textbf{99.16} & 65.70 & 51.11 & 55.17 & 50.04 & 98.05 & 92.65 & 98.06  & \textbf{94.26} & 68.42 & 54.30 & 85.90 & 74.26 & 86.11 & 77.14 & 86.59  & 78.35 \\ 
            \hline
            + Grad-AAT & 98.91 & 98.10 & 79.47 & 58.48 & 58.74 & 52.96 & 98.07 & 92.74 & 97.71 & 93.42 & 70.86 & 57.33 & 87.25 & 76.22 & 86.07 & 77.95 & 86.10 & 78.08 \\ 
            \hline
            + Grad-SAT & 98.72 & 98.03 & 73.59 & 56.93 & 57.77 & 52.09 & 97.98 & 92.61 & 97.56 & 93.33 & 69.68 & 55.48 & 86.14 & 75.15 & 86.08 & 77.20 & 85.87 & 77.96 \\ 
            \hline
            + Grad-BAT & 99.27 & 98.36 & 90.40 & 67.45 & 59.72 & 54.10 & 98.38 & 93.13 & 98.40 & 93.80 & 72.41 & 58.54 & 87.62 & 76.53 & 86.39 & 78.46 & 86.18 & 78.44 \\ 
            \hline
            + Gen-BAT & 97.75 & 97.31 & 94.63 & 79.49 & 68.28 & 56.41 & \textbf{98.46} & 93.34 & 98.44 & 93.91 & 72.85 & 58.99 & 88.41 & 77.96 & 87.32 & 79.93 & 86.15 & 78.32 \\ 
            \hline
            + Two-Gen-BAT & 97.86 & 97.66 & \textbf{95.28} & \textbf{81.34} & \textbf{69.97} & \textbf{59.27} & 98.42 & 93.35 & 98.61 & 94.15 & \textbf{73.40} & 59.40 & 88.78 & \textbf{79.45} & \textbf{90.77} & 83.39 & \textbf{89.50} & \textbf{81.80} \\ 
            \hline
            \hline
            + Combined AT & 98.56 & 98.20 & 95.14 & 81.30 & 69.66 & 58.61 & 98.40 & \textbf{93.46} & \textbf{98.68} & 93.99 & 72.87 & \textbf{59.76} & \textbf{88.91} & 79.42 & \textbf{90.77} & \textbf{83.45} & 88.52 & 80.75 \\ 
            \hline
        \end{tabular}
    \end{center}
\end{table*}

\begin{table*}[!htbp]
    \setlength\tabcolsep{5.0pt} 
        \footnotesize
      \caption{Comparison between adversarial training and traditional data augmentation in improving generalization to unseen face forgery technologies (based on Xception~\cite{xception}). The models were trained using the NT C23 data in FF++, and tested on data generated using different forgery technologies (indicated by ``NT $\rightarrow\ast$'').} 
      \vskip -0.15in
      \label{table:sup aug type}
      \begin{center}
        \begin{tabular}{c|c|c|c|c|c|c|c|c|c|c|c|c|c|c}
            \hline 
            Method & \multicolumn{12}{c}{NT $\rightarrow\ast$} \\
            \hline
            & \multicolumn{2}{c|}{NT} & \multicolumn{2}{c|}{DF} & \multicolumn{2}{c|}{F2F} & \multicolumn{2}{c|}{FS} & \multicolumn{2}{c|}{DFD} & \multicolumn{2}{c|}{Celeb-DF} & \multicolumn{2}{c}{Avg} \\
            \hline
            & AUC & ACC & AUC & ACC & AUC & ACC & AUC & ACC & AUC & ACC & AUC & ACC & AUC & ACC \\
            & (\%) & (\%) & (\%) & (\%) & (\%) & (\%) & (\%) & (\%) & (\%) & (\%) & (\%) & (\%) & (\%) & (\%) \\
            \hline 
            Xception~\cite{xception} & 98.06 & 94.26 & 82.66 & 60.54 & 60.96 & 51.07 & 42.94 & 48.37 & 69.82 &  --  & 53.90 & 53.75 & 68.06 & 61.60\\
            \hline 
            + Gaussian Noise & 98.53 & 94.58 & 82.75 & 61.60 & 69.96 & 51.16 & 42.99 & 48.46 & 69.75 & -- &  55.01 & 53.73 & 69.85 & 61.91 \\
            \hline 
            + Gaussian Blur & 98.87 & 95.05 & 82.75 & 61.64 & 70.02 & 51.20 & 43.02 & 48.52 & 69.95 &  --  & 54.98 & 54.68 & 69.93 & 62.22 \\
            \hline 
            + JPEG Compression & 98.44 & 94.53 & 82.69 & 61.57 & 70.00 & 51.14 & 42.98 & 48.42 & 69.88 &  --  & 55.10 & 53.93 & 69.85 & 61.92 \\
            \hline 
            + Combined Traditional & \textbf{98.99} & \textbf{95.08} & 82.77 & 61.70 & 70.07 & 51.28 & 43.10 & 48.53 & 70.03 & -- & 55.16 & 54.79 & 70.02 & 62.28\\
            \hline             
            + Two-Gen-BAT (ours) & 98.61 & 94.15 & 86.18 & 64.54 & 72.02 & 55.35 & 45.87 & 50.47 & 73.15 &  --  & 59.69 & 60.29 & 72.59 & 65.16\\
            \hline 
            + Combined AT (ours) & 98.68 & 93.99 & \textbf{86.58} & \textbf{65.42} & \textbf{72.75} & \textbf{55.58} & \textbf{46.16} & \textbf{50.80} & \textbf{74.45} & -- & \textbf{61.47} & \textbf{62.06} & \textbf{73.35} & \textbf{65.57}\\
            \hline
        \end{tabular}
    \end{center}
\end{table*}

\begin{table*}[!htbp]
    \setlength\tabcolsep{2.0pt} 
        \footnotesize
      \caption{Comparison between adversarial training and traditional data augmentation in improving generalization to unseen image/video qualities (based on Xception~\cite{xception}). The train and test data were split from the NT data in FF++~\cite{rossler2019faceforensics++}, of possibly different qualities. The $\rightarrow$ symbol indicates the models were trained on a specific image/video quality, and were expected to generalize to other image/video qualities.} 
      \vskip -0.15in
      \label{table:sup aug quality}
      \begin{center}
        \begin{tabular}{c|c|c|c|c|c|c|c|c|c|c|c|c|c|c|c|c|c|c}
            \hline
            Method & \multicolumn{6}{c|}{Raw $\rightarrow\ast$} & \multicolumn{6}{c|}{C23 $\rightarrow\ast$} & \multicolumn{6}{c}{C40 $\rightarrow\ast$} \\
            \hline
             & \multicolumn{2}{c|}{Raw} & \multicolumn{2}{c|}{C23} & \multicolumn{2}{c|}{C40} & \multicolumn{2}{c|}{Raw} & \multicolumn{2}{c|}{C23} & \multicolumn{2}{c|}{C40} & \multicolumn{2}{c|}{Raw} & \multicolumn{2}{c|}{C23} & \multicolumn{2}{c}{C40}\\
            \hline
            & AUC & ACC & AUC & ACC & AUC & ACC & AUC & ACC & AUC & ACC & AUC & ACC & AUC & ACC & AUC & ACC & AUC & ACC \\
            & (\%) & (\%) & (\%) & (\%) & (\%) & (\%) & (\%) & (\%) & (\%) & (\%) & (\%) & (\%) & (\%) & (\%) & (\%) & (\%) & (\%) & (\%) \\
            \hline 
            Xception~\cite{xception} & 99.21  & 99.16 & 65.70 & 51.11 & 55.17 & 50.04 & 98.05 & 92.65 & 98.06  & 94.26 & 68.42 & 54.30 & 85.90 & 74.26 & 86.11 & 77.14 & 86.59  & 78.35 \\ 
            \hline
            + Gaussian Noise & 98.79 & 98.77 & 74.09 & 56.18 & 56.95 & 51.86 & 97.82 & 92.59 & 98.53 & 94.58 & 70.68 & 59.82 & 87.38 & 75.45 & 86.44 & 78.53 & 86.39 & 78.35\\ 
            \hline
            + Gaussian Blur & \textbf{99.26} & 99.14 & 78.42 & 59.65 & 57.59 & 52.34 & 98.15 & 92.98 & 98.87 & 95.05 & 70.89 & 59.98 & 87.58 & 75.69 & 86.83 & 78.76 & 86.44 & 78.45 \\ 
            \hline
            + JPEG Compression & 98.92 & 98.88 & 83.86 & 67.76 & 66.24 & 54.35 & 97.90 & 92.69 & 98.44 & 94.53 & 71.96 & 61.27 & 87.45 & 76.33 & 87.01 & 79.39 & 87.53 & 79.56 \\ 
            \hline
            + Combined Traditional & 99.14 & \textbf{99.18} & 84.85 & 68.90 & 66.75 & 55.38 & \textbf{98.51} & 92.97 & \textbf{98.99} & \textbf{95.08} & 72.16 & \textbf{62.03} & 87.57 & 76.52 & 87.45 & 80.40 & 87.57 & 79.06 \\ 
            \hline
            + Two-Gen-BAT (ours) & 97.86 & 97.66 & \textbf{95.28} & \textbf{81.34} & \textbf{69.97} & \textbf{59.27} & 98.42 & 93.35 & 98.61 & 94.15 & \textbf{73.40} & 59.40 & 88.78 & \textbf{79.45} & \textbf{90.77} & 83.39 & \textbf{89.50} & \textbf{81.80} \\ 
            \hline
            + Combined AT (ours) & 98.56 & 98.20 & 95.14 & 81.30 & 69.66 & 58.61 & 98.40 & \textbf{93.46} & 98.68 & 93.99 & 72.87 & 59.76 & \textbf{88.91} & 79.42 & \textbf{90.77} & \textbf{83.45} & 88.52 & 80.75 \\ 
            \hline
        \end{tabular}
    \end{center}
\end{table*}

\begin{table}[h!]
    \setlength\tabcolsep{1.5pt} 
      \footnotesize
      \caption{Adversarial accuracy under different kernel size. }
      \label{table:attack kernel size}
      \begin{center}\vskip -0.1in
        \begin{tabular}{p{1.5cm}<{\centering}|p{1.5cm}<{\centering}|p{1.5cm}<{\centering}|p{1.5cm}<{\centering}|p{1.5cm}<{\centering}}
            \hline 
            \multicolumn{2}{c|}{} & \multicolumn{3}{c}{Target} \\
            \hline
           Kernel Size & Source & Inc-v3 & Inc-v4 & IncRes-v2 \\
            \hline
           \multirow{3}{*}{3} & Inc-v3 & 29.08\% & 75.60\% & 76.68\%\\
            & Inc-v4 & 72.30\% & 38.78\% & 76.28\%\\
            & IncRes-v2 & 72.30\% & 75.60\% & 39.92\%\\
           \hline
           \multirow{3}{*}{5} & Inc-v3 & 21.48\% & 71.04\% & 71.18\%\\
            & Inc-v4 & 65.86\% & 30.84\% & 71.18\%\\
            & IncRes-v2 & 65.84\% & 71.04\% & 33.20\%\\
           \hline
           \multirow{3}{*}{7} & Inc-v3 & 16.80\% & 65.68\% & 65.98\%\\
            & Inc-v4 & 60.12\% & 23.04\% & 65.98\%\\
            & IncRes-v2 & 60.12\% & 65.52\% & 27.28\%\\
           \hline
           \multirow{3}{*}{9} & Inc-v3 & 15.88\% & 62.30\% & 61.92\%\\
            & Inc-v4 & 56.14\% & 20.92\% & 62.02\%\\
            & IncRes-v2 & 56.22\% & 62.00\% & 25.62\%\\
           \hline
        \end{tabular}
    \end{center}
\end{table}

Table~\ref{table:attack kernel size} summarizes how the kernel size of adversarial blurring affects the attack performance. We keep using the adversarial accuracy (\ie, the prediction accuracy of the victim model on adversarial examples) as an evaluation metric. 
Figure~\ref{fig: kernel} demonstrates the adversarial examples generated using different kernel sizes. We let $\epsilon$ be $0.1$ and set $\sigma$ be a map of all ones for Table~\ref{table:attack kernel size} and Figure~\ref{fig: kernel}.
As expected, as the kernel size increases, we can obtain more blurry images and thus achieve higher attack success rates.

Table~\ref{table:attack sigma} summarizes the influence of initialization of the $\sigma^{adv}$ map (\ie, by settings different $\sigma$ maps) for performing the adversarial blurring attack. We let $\epsilon$ be 0.1 and fixed the kernel size to $5$ for obtaining results in Table~\ref{table:attack sigma}.

\begin{table}[htbp]
    \setlength\tabcolsep{1.5pt} 
      \footnotesize
      \caption{Adversarial accuracy of ImageNet~\cite{imagenet} models under different initialization of $\sigma^{adv}$. }\vskip -0.15in
      \label{table:attack sigma}
      \begin{center}
        \begin{tabular}{p{1.5cm}<{\centering}|p{1.5cm}<{\centering}|p{1.5cm}<{\centering}|p{1.5cm}<{\centering}|p{1.5cm}<{\centering}}
            \hline 
            \multicolumn{2}{c|}{} & \multicolumn{3}{c}{Target} \\
             \hline
            Initial value & Source & Inc-v3 & Inc-v4 & IncRes-v2 \\
             \hline
             \multirow{3}{*}{0.1} & Inc-v3 & 49.08\% & 76.06\% & 77.28\%\\
             & Inc-v4 & 75.10\% & 54.68\% & 78.28\%\\
             & IncRes-v2 & 75.10\% & 78.06\% & 58.38\%\\
            \hline
            \multirow{3}{*}{1} & Inc-v3 & 21.48\% & 71.04\% & 71.18\%\\
            & Inc-v4 & 65.86\% & 30.84\% & 71.18\%\\
            & IncRes-v2 & 65.84\% & 71.04\% & 33.20\%\\
           \hline
             \multirow{3}{*}{10} & Inc-v3 & 18.12\% & 69.96\% & 70.24\%\\
             & Inc-v4 & 63.76\% & 27.58\% & 69.24\%\\
             & IncRes-v2 & 63.78\% & 69.98\% & 29.30\%\\
            \hline
             \multirow{3}{*}{100} & Inc-v3 & 17.20\% & 68.78\% & 69.24\%\\
             & Inc-v4 & 62.86\% & 25.64\% & 68.22\%\\
             & IncRes-v2 & 62.84\% & 68.18\% & 28.70\%\\
            \hline
          
        \end{tabular}
    \end{center}
\end{table}

\begin{table}[!htbp]
    \setlength\tabcolsep{3.0pt} 
        \footnotesize
      \caption{Adversarial accuracy of ImageNet models on our adversarial examples and the PGD examples. We choose the iteration number to be 3}
      \vskip -0.15in
      \label{table:iter attack}
      \begin{center}
        \begin{tabular}{c|c|c|c|c|c|c}
            \hline 
            Source & \multicolumn{6}{c}{Target} \\
             \hline
             & \multicolumn{2}{c|}{Inc-v3} & \multicolumn{2}{c|}{Inc-v4} & \multicolumn{2}{c}{IncRes-v2}\\
            \hline
             & PGD & BAT & PGD & BAT & PGD & BAT\\
            \hline
            Inc-v3 & 22.06\% & 18.53\% & 69.73\% & 69.02\% & 69.39\% & 68.26\% \\
            \hline 
            Inc-v4 & 67.63\% & 62.95\% & 33.84\% & 28.89\% & 68.84\% & 68.06\% \\
            \hline
            IncRes-v2 & 65.92\% & 63.13\% & 68.29\% & 67.72\% & 41.84\% & 31.74\%\\
            \hline
        \end{tabular}
    \end{center}
\end{table}

\section{Multi-step Scheme in Adversarial Training}\label{sec:sup multi}
As has been mentioned in the main paper, multi-step adversarial examples can be introduced in the input-gradient-based adversarial training, though the training complexity shall increase drastically. We report some experimental results in the multi-step scheme. We tested the performance of using multi-step FGSM (or called iterative FGSM~\cite{kurakin2016adversarial}) and multi-step (input-gradient-based) adversarial blurring in training deepfake detection models. We considered three steps in this experiment, and it was performed based on the EfficientNet backbone. Our results showed that the three-step FGSM 
in general did not lead to superior test set AUCs (NT: 98.61\%, DF: 85.12\%, F2F: 68.05\%, FS: 44.58\%) in comparison to the single-step FGSM (\ie, Grad-AAT, NT: 98.00\%, DF: 85.91\%, F2F: 71.12\%, FS: 44.97\%). For our adversarial blurring, the three-step scheme seems slightly more effective in generalizing to the DF and FS data (NT: 98.75\%, DF: 86.48\%, F2F: 69.67\%, FS: 46.72\%) in comparison to the single-step input-gradient-based scheme, yet further increasing the number of steps issued in similar or worse AUCs. 

The training complexity of our two-generator-based BAT is similar to that of the three-step input-gradient-based BAT, showing that the generator-based method well trades off the test-set accuracy against the training complexity. 

\section{The Role of Kernel Size in Adversarial Training}\label{sec:sup kernel-size}
In order to better explore the impact of kernel size on our approach, we conducted a set of experiments with only different kernel sizes, and compare the performance of obtained models.
It can be seen from Table \ref{table:ablation kernel size for type} and \ref{table:ablation kernel size for quality} that when the kernel size is increased, the generalization performance of the model is also improved. Future work can consider even larger kernel sizes. 

\begin{table*}[htb]
    \setlength\tabcolsep{5.0pt} 
        \footnotesize
      \caption{Comparison between Two-Gen-BAT settings with different kernel sizes in improving the generalization to unseen forgery technologies. Except for the kernel size, all configurations are consistent with Table \ref{table:ablation type}.}
      \vskip -0.15in
      \label{table:ablation kernel size for type}
      \begin{center}
        \begin{tabular}{c|c|c|c|c|c|c|c|c|c|c|c|c|c|c}
            \hline 
             & \multicolumn{12}{c}{NT $\rightarrow\ast$} \\
            \hline
            Kernel Size & \multicolumn{2}{c|}{NT} & \multicolumn{2}{c|}{DF} & \multicolumn{2}{c|}{F2F} & \multicolumn{2}{c|}{FS} & \multicolumn{2}{c|}{DFD} & \multicolumn{2}{c|}{Celeb-DF} & \multicolumn{2}{c}{Avg} \\
            \hline
            & AUC & ACC & AUC & ACC & AUC & ACC & AUC & ACC & AUC & ACC & AUC & ACC & AUC & ACC \\
            & (\%) & (\%) & (\%) & (\%) & (\%) & (\%) & (\%) & (\%) & (\%) & (\%) & (\%) & (\%) & (\%) & (\%) \\
            \hline 
            3 & 98.11 & 95.00 & 86.50 & 66.92 & 71.31 & 55.05 & 46.98 & 49.34 & 76.05 &  --  & 65.43 & 66.94 & 74.06 & 66.65\\
            \hline 
            5 & 98.37 & 95.14 & 87.04 & 68.01 & 72.83 & 55.48 & 47.72 & 49.87 & 76.26 &  --  & 66.29 & 67.39 & 74.75 & 67.18\\
            \hline 
            7 & 98.54 & 95.21 & 87.38 & 68.93 & 73.77 & 55.86 & 48.52 & 50.12 & 76.43 &  --  & 66.58 & 67.74 & 75.20 & 67.57\\
            \hline 
            9 & \textbf{98.72} & \textbf{95.26} & \textbf{87.51} & \textbf{69.40} & \textbf{74.65} & \textbf{56.19} & \textbf{48.99} & \textbf{50.43} & \textbf{76.60} & -- & \textbf{66.84} & \textbf{67.91} & \textbf{75.55} & \textbf{67.84}\\
            \hline 

        \end{tabular}
    \end{center}
\end{table*}

\begin{table*}[htb]
    \setlength\tabcolsep{2.0pt} 
        \footnotesize
      \caption{Comparison between Two-Gen-BAT settings with different kernel sizes in improving the generalization to unseen image/video qualities (indicated by $\rightarrow$). Except for the kernel size, all configurations are consistent with Table \ref{table:ablation quality}.} 
      \vskip -0.15in
      \label{table:ablation kernel size for quality}
      \begin{center}
        \begin{tabular}{c|c|c|c|c|c|c|c|c|c|c|c|c|c|c|c|c|c|c}
            \hline
             & \multicolumn{6}{c|}{Raw $\rightarrow\ast$} & \multicolumn{6}{c|}{C23 $\rightarrow\ast$} & \multicolumn{6}{c}{C40 $\rightarrow\ast$} \\
            \hline
            Kernel Size & \multicolumn{2}{c|}{Raw} & \multicolumn{2}{c|}{C23} & \multicolumn{2}{c|}{C40} & \multicolumn{2}{c|}{Raw} & \multicolumn{2}{c|}{C23} & \multicolumn{2}{c|}{C40} & \multicolumn{2}{c|}{Raw} & \multicolumn{2}{c|}{C23} & \multicolumn{2}{c}{C40}\\
            \hline
            & AUC & ACC & AUC & ACC & AUC & ACC & AUC & ACC & AUC & ACC & AUC & ACC & AUC & ACC & AUC & ACC & AUC & ACC \\
            & (\%) & (\%) & (\%) & (\%) & (\%) & (\%) & (\%) & (\%) & (\%) & (\%) & (\%) & (\%) & (\%) & (\%) & (\%) & (\%) & (\%) & (\%) \\
            \hline 
            3 & 99.11 & 98.20 & 95.75 & 82.13 & 69.79 & 59.14 & 98.64 & 94.70 & 98.11 & 95.00 & 73.42 & 61.91 & 90.02 & 82.01 & 92.48 & 84.43 & 89.02 & 80.93 \\ 
            \hline
            5 & 99.39 & 98.64 & 95.84 & 83.07 & 70.74 & 60.26 & 98.82 & 94.98 & 98.37 & 95.14 & 73.86 & 62.20 & 90.32 & 82.43 & 92.81 & 84.77 & 91.13 & 83.04 \\ 
            \hline
            7 & 99.43 & 98.89 & 95.89 & 83.81 & 71.55 & 61.05 & \textbf{98.95} & 95.36 & 98.54 & 95.21 & 74.21 & 62.45 & 90.46 & 82.74 & 93.29 & 85.04 & 92.84 & 85.46 \\ 
            \hline
            9 & \textbf{99.46} & \textbf{99.04} & \textbf{95.93} & \textbf{84.19} & \textbf{72.18} & \textbf{61.87} & 98.92 & \textbf{95.52} & \textbf{98.72} & \textbf{95.26} & \textbf{74.73} & \textbf{62.53} & \textbf{90.58} & \textbf{82.91} & \textbf{93.51} & \textbf{85.36} & \textbf{94.19} & \textbf{87.02} \\ 
            \hline
        \end{tabular}
    \end{center}
\end{table*}

\section{Hyper-parameter Tuning}\label{sec:sup hyper}
In order to compare the traditional data augmentation methods and our method fairly, we fine-tuned each hyper-parameter carefully on the validation set and report test results using those obtained the best validation accuracies. 
For data augmentation using the traditional Gaussian noise, we fine-tuned the following hyper-parameters: $p\in[0, 1.0]$, \ie, the probability of adding noise, and the mean and variance of the noise, in the range of $[-1.0,1.0]$ and $[0, 50]$, respectively. 
We saw the best validation performance when using $p=0.5$, $\mu=0$, and an uniformly random variance in the range of $[0,30]$ for each training image.
Testing with the traditional Gaussian blurring augmentation, we set the kernel size to 9, and we fine-tuned the variance of each Gaussian kernel and also $p\in[0, 1.0]$. 
In this setting, the best validation performance was obtained when each training image took a uniformly random variance in the range of $[0,30]$. 
For JPEG compression, we let the compression quality for each image be randomly sampled between a lower bound and an upper bound. Empirical results on the validation set suggested that the two bounds be $60$ and $100$, respectively. 
For the combination of traditional augmentations, we took the same values for all these hyper-parameters.

\section*{Acknowledgements}
This work was partially supported by National Key RD Program of China under Grant 2021ZD0112100 and National Natural Science Foundation of China under Grants No. U19A2073.

%

\ifCLASSOPTIONcaptionsoff
  \newpage
\fi



\bibliographystyle{IEEEtran}
\bibliography{IEEEabrv,egbib}
%



%

\begin{IEEEbiography}[{\includegraphics[width=1in,height=1.25in,clip,keepaspectratio]{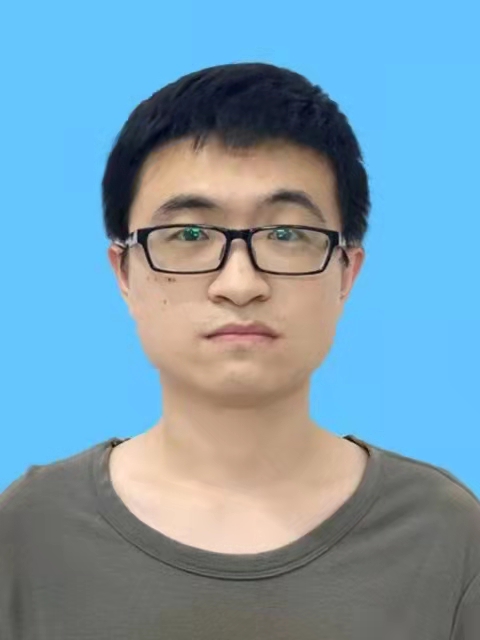}}]{Zhi Wang}
received the M.Sc. degree in computer science and technology from the Harbin Institute of Technology, Harbin, China, in 2021. His research interests include deep learning
and computer vision.
\end{IEEEbiography}

\begin{IEEEbiography}[{\includegraphics[width=1in,height=1.25in,clip,keepaspectratio]{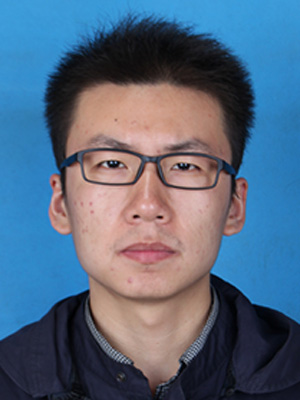}}]{Yiwen Guo} 
received the B.E. degree from Wuhan University, Wuhan, China, in 2011, and the Ph.D. degree from Tsinghua University, Beijing, China, in 2016. He is an independent researcher. Prior to this, he was a research scientist at ByteDance AI Lab and a staff research scientist at Intel Labs. His current research interests include computer vision, pattern recognition, and machine learning.
\end{IEEEbiography}

\begin{IEEEbiography}[{\includegraphics[width=1in,height=1.25in,clip,keepaspectratio]{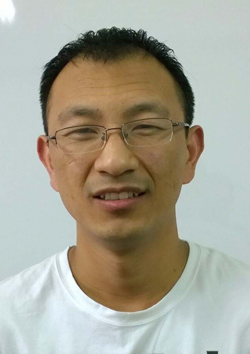}}]{Wangmeng Zuo} (M'09, SM'14)  received the Ph.D. degree in computer application technology from the Harbin Institute of Technology, Harbin, China, in 2007.
He is currently a Professor in the School of Computer Science and Technology, Harbin Institute of Technology. His current research interests include image enhancement and restoration, image and face editing, object detection, visual tracking, and image classification. He has published over 100 papers in top tier academic journals and conferences. He has served as a Tutorial Organizer in ECCV 2016, an Associate Editor of \emph{IEEE Trans. Pattern Analysis and Machine Intelligence}, \emph{The Visual Computers}, \emph{Journal of Electronic Imaging}.
\end{IEEEbiography}





\end{document}